%% file: main.tex
\documentclass[10pt,journal,compsoc]{IEEEtran}
\ifCLASSOPTIONcompsoc
  \usepackage[nocompress]{cite}
\else
  \usepackage{cite}
\fi
\ifCLASSINFOpdf
\else
\fi
\usepackage{url}
\usepackage{amsmath} 
\usepackage{amssymb}
\usepackage{graphicx}
\usepackage{booktabs}
\usepackage{cite}
\usepackage{forest}

\begin{document}
%
\title{A Survey of RWKV}
%
%
%
%

\author{Zhiyuan Li , Tingyu Xia, Yi Chang and Yuan Wu
\IEEEcompsocitemizethanks{
\IEEEcompsocthanksitem Zhiyuan Li and Tingyu Xia are equal contributions.\protect\\
E-mail: zhiyuanl24@mails.jlu.edu.cn, xiaty21@mails.jlu.edu.cn.
\IEEEcompsocthanksitem Yuan Wu is the corresponding author.\protect\\
E-mail: yuanwu@jlu.edu.cn.
\IEEEcompsocthanksitem All authors  are with School of  Artificial Intelligence, Jilin University.
\IEEEcompsocthanksitem Yi Chang is with Engineering Research Center of Knowledge-Driven Human-Machine Intelligence, MOE, China\protect\\
E-mail: yichang@jlu.edu.cn.}
}

%
%

\markboth{Journal of \LaTeX\ Class Files,~Vol.~14, No.~8, August~2015}%
{Shell \MakeLowercase{\textit{et al.}}: Bare Demo of IEEEtran.cls for Computer Society Journals}
%



\IEEEtitleabstractindextext{%
\input{tex/abstract}
}

\maketitle

\IEEEdisplaynontitleabstractindextext

%
\IEEEpeerreviewmaketitle

\input{tex/introduction}

\input{tex/background}

\input{tex/rwkv}

\input{tex/applications}

\input{tex/future}

\section{Conclusion}
The RWKV model, an innovative deep learning framework, has demonstrated significant potential across numerous fields, such as natural language processing, time series analysis, and computer vision. Its distinct structure achieves an optimal blend of high-level performance and computational efficiency, making it apt for tackling intricate tasks. However, even with its increasing influence, there is a noticeable absence of an exhaustive review of the RWKV model and its applications. This paper aims to address this deficiency by delivering an in-depth analysis of the RWKV model, emphasizing its architectural advancements, primary applications, and the fields where it has shown notable success. Furthermore, we examine the existing challenges in RWKV-related research and suggest promising avenues for future investigation.


%





\ifCLASSOPTIONcaptionsoff
  \newpage
\fi



%

\bibliographystyle{IEEEtran} 
\bibliography{ref}




%





\end{document}

%% file: tex/abstract.tex
\begin{abstract}
The Receptance Weighted Key Value (RWKV) model offers a novel alternative to the Transformer architecture, merging the benefits of recurrent and attention-based systems. Unlike conventional Transformers, which depend heavily on self-attention, RWKV adeptly captures long-range dependencies with minimal computational demands. By utilizing a recurrent framework, RWKV addresses some computational inefficiencies found in Transformers, particularly in tasks with long sequences. RWKV has recently drawn considerable attention for its robust performance across multiple domains.  Despite its growing popularity, no systematic review of the RWKV model exists. This paper seeks to fill this gap as the first comprehensive review of the RWKV architecture, its core principles, and its varied applications, such as natural language generation, natural language understanding, and computer vision. We assess how RWKV compares to traditional Transformer models, highlighting its capability to manage long sequences efficiently and lower computational costs. Furthermore, we explore the challenges RWKV encounters and propose potential directions for future research and advancement. We consistently maintain the related open-source materials at: \url{https://github.com/MLGroupJLU/RWKV-Survey}.
\end{abstract}

\begin{IEEEkeywords}
RWKV, Large Language Model, Foundation Models, Deep Learning.
\end{IEEEkeywords}

%% file: tex/introduction.tex
\IEEEraisesectionheading{\section{Introduction}\label{sec:introduction}}

\IEEEPARstart{A}{s} an important branch of machine learning, deep learning originated from the perceptron model~\cite{rosenblatt1958perceptron}, which laid the foundation for the development of neural networks. Over the past decade, deep learning has rapidly advanced and been widely applied, profoundly impacting various fields.  Classic deep learning models, including Multilayer Perceptron (MLP), Convolutional Neural Networks (CNN), and Recurrent Neural Networks (RNN), have achieved significant success in areas such as image recognition~\cite{lecun1998gradient, zhu2020deep, touvron2022resmlp}, speech recognition~\cite{hinton2012deep, dahl2011context, zhang2020transformer}, and natural language processing~\cite{cambria2014jumping, incitti2023beyond}.  For example, CNNs have become central to computer vision~\cite{krizhevsky2017imagenet,he2016deep,wu2020dual}, with applications ranging from facial recognition to automatic driving~\cite{badue2021self,raji2021face,yurtsever2020survey}. Meanwhile, RNNs and their variants, particularly long-short-term memory (LSTM) networks~\cite{hochreiter1997long}, have proven essential for natural language processing tasks, especially in speech recognition and machine translation~\cite{bahdanau2014neural,dabre2020survey,malik2021automatic}.
 
Proposed by Vaswani et al. in 2017, the Transformer model~\cite{vaswani2017attention} has since become a dominant architecture in natural language processing (NLP)~\cite{roy2021efficient,ouyang2022training,fedus2022switch} and various other fields~\cite{zhou2022ibot,ren2022sigt,zhang2022bigssl}. Transformer-based models such as BERT~\cite{devlin2018bert} and GPT~\cite{radford2018improving} have demonstrated exceptional performance in tasks such as text classification~\cite{tezgider2022text}, language generation~\cite{zhang2023survey}, and question answering~\cite{shao2019transformer}. The Transformer utilizes a self-attention mechanism to adeptly capture long-distance dependencies in input sequences, providing a more robust approach for modeling intricate data relationships than earlier models like RNNs and LSTMs~\cite{dai2019transformer}. Furthermore, the parallelization of both training and inference processes enhances the scalability of the Transformer, significantly improving processing speed and model performance. However, despite these advantages, the Transformer faces significant challenges. The computational cost of the attention mechanism grows quadratically with the sequence length, leading to inefficient memory usage and slower processing of long sequences~\cite{dong2021attention}. This issue becomes particularly problematic in tasks that require handling large amounts of data. Despite numerous endeavors to explore methods for reducing the complexity of the Transformer model, unfortunately, significant progress in substantially decreasing its complexity has yet to be achieved~\cite{choromanski2020rethinking, katharopoulos2020transformers, wang2020linformer}.

To address the limitations of the Transformer model, recent advancements have introduced state-space models~\cite{gu2021efficiently}, which provide a promising solution for efficiently handling long sequences while maintaining strong contextual understanding. Models such as Mamba~\cite{gu2023mamba} and RetNet~\cite{sun2023retentive} have been proposed to reduce computational complexity in sequence data processing. Mamba aims to enhance the efficiency of long-sequence processing by modifying the attention mechanism and reducing memory usage in the process. Similarly, RetNet employs a recurrent structure to improve memory retention and extend the model’s ability to capture long-range dependencies, all while avoiding the quadratic time complexity that affects traditional Transformers. One of the most notable innovations in recent times is the Receptance Weighted Key Value (RWKV) model~\cite{peng2023rwkv}, which effectively combines the strengths of RNNs with the attention-based mechanism of Transformers. The RWKV model utilizes a unique key-value approach to manage dependencies sequentially, significantly reducing computational overhead while preserving the ability to model long-term dependencies with minimal memory requirements. RWKV has already demonstrated its potential across a wide range of applications, from NLP tasks such as text generation~\cite{AI-Writer}, machine translation~\cite{liu2023approach}, and sentiment analysis to time series prediction tasks~\cite{hao2024multi}.

\begin{figure*}[!htbp]
	\centering
	\resizebox{\linewidth}{!}{
	\input{Figures/fig-tree}
	}
	\caption{Structure of this paper.}
	\label{fig-main}
\end{figure*}
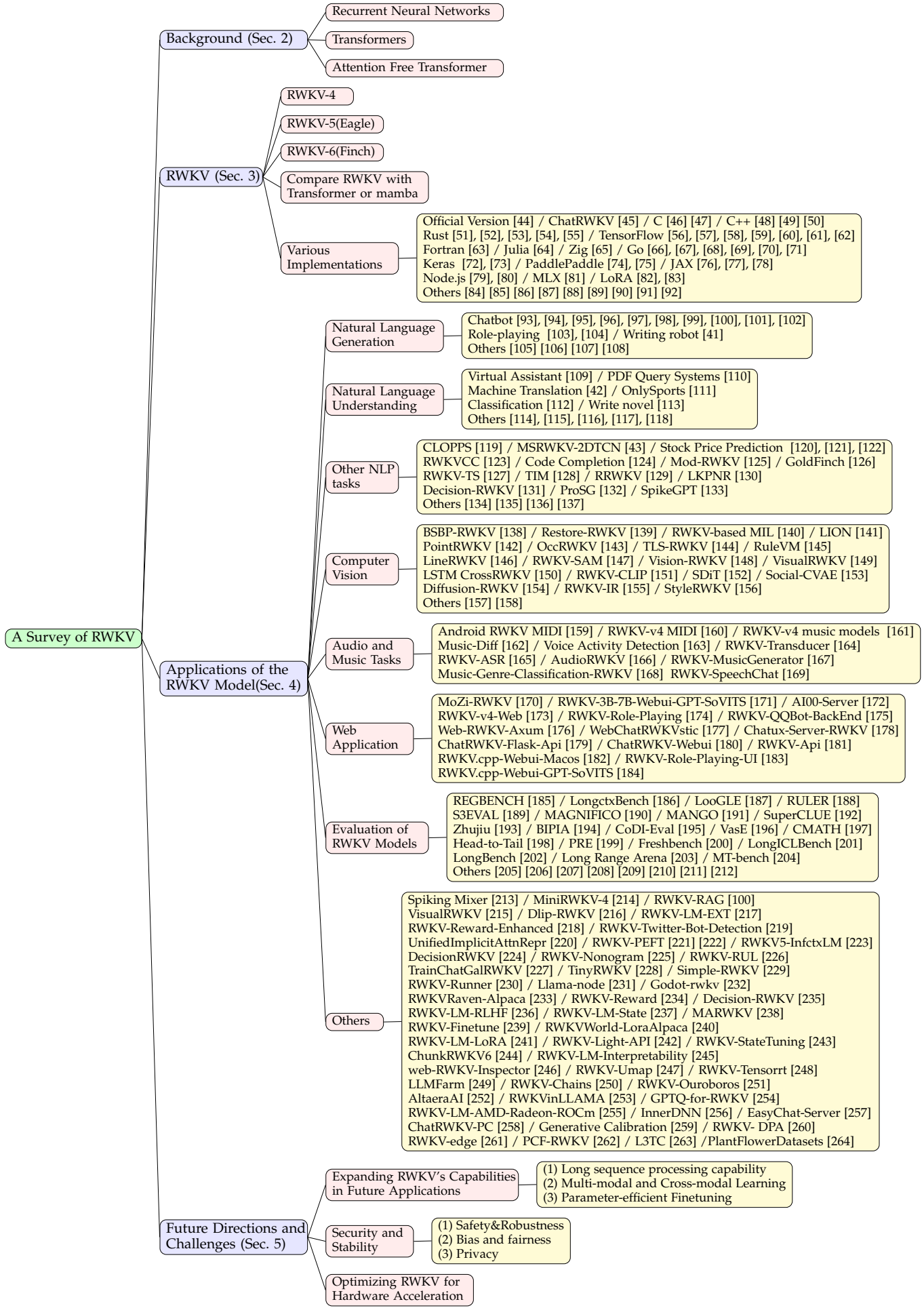

Inspired by the exceptional efficiency of RWKV in handling long sequences and its growing influence in various fields, there has been a surge in research exploring its capabilities and potential applications. Given the increasing interest in RWKV, this paper aims to provide a comprehensive overview of the model, its evolution, and future directions. As depicted in \figurename~\ref{fig-main}, the remainder of this paper is organized as follows: Section 2 presents a systematic review of foundational concepts, covering RNNs, Transformer architectures, and Attention-Free Transformer (AFT). Section 3 delves into the principles and implementations of RWKV, providing a comparative analysis with Transformer and Mamba, along with a detailed discussion of its various implementations. Section 4 thoroughly explores various applications of RWKV, including Natural Language Generation (NLG), Natural Language Understanding (NLU), other NLP tasks, computer vision, web applications, and evaluation methods. Finally, Section 5 provides the key challenges RWKV faces and its future research directions.

%% file: Figures/fig-tree.tex
\begin{forest}
  for tree={
  grow=east,
  reversed=true,
  anchor=base west,
  parent anchor=east,
  child anchor=west,
  base=left,
  font=\small,
  rectangle,
  draw,
  rounded corners,align=left,
  minimum width=2.5em,
  inner xsep=4pt,
  inner ysep=1pt,
  },
  where level=1{text width=5em,fill=blue!10}{},
  where level=2{text width=5em,font=\footnotesize,fill=pink!30}{},
  where level=3{font=\footnotesize,yshift=0.26pt,fill=yellow!20}{},
  [A Survey of RWKV,fill=green!20
        [Background (Sec.~\ref{sec:background}),text width=9em
            [Recurrent Neural Networks,text width=11em]
            [Transformers,text width=5em]
            [Attention Free Transformer,text width=11em]
        ]
        [RWKV (Sec.~\ref{sec:rwkv}),text width=6em
            [RWKV-4,text width=4em]
            [RWKV-5(Eagle),text width=6em]
            [RWKV-6(Finch),text width=6em]
            [Compare RWKV with \\ Transformer or mamba,text width=9em]
            [Various \\ Implementations,text width=7em 
            [Official Version~\cite{PENG_RWKV-LM_2021} / ChatRWKV~\cite{ChatRWKV} / C~\cite{rwkv.cpp}~\cite{rwkv.c} / C++~\cite{rwkv-cpp-accelerated}~\cite{rwkv-cpp-server}~\cite{rwkv_tokenizer_cpp}  \\ Rust~\cite{web-rwkv, rwkvk-rs, rwkv-tokenizer, smolrsrwkv, rwkv-rust}  / TensorFlow~\cite{rwkv-tensorflow, RWKV5-Tensorflow2.0, RWKV-Tensorflow2.0, Tensor-RWKV, rwkv-cuda, tensorflow-rwkv, RWKV-tf2}  \\ Fortran~\cite{rwkv.f90} / Julia~\cite{rwkv.jl} / Zig~\cite{rwkv.zig} / Go~\cite{rwkv, rwkv_go, go-rwkv.cpp, verbaflow, rwkvalgorithm, rwkv-tokenizer-go} \\ Keras ~\cite{RWKV6-Keras, keras-rwkv} / PaddlePaddle~\cite{rwkv-paddle, RWKV-v2-RNN-paddle} / JAX~\cite{tpu_rwkv, RWKV-LM-jax, rwkv-jax}  \\ Node.js~\cite{RWKV-cpp-node, RWKV-tokenizer-node} / MLX~\cite{mlx-rwkv} / LoRA~\cite{RWKV5-LM-LoRA, RWKV-infctx-trainer-LoRA} \\ Others~\cite{RWKV-infctx-trainer}  \cite{RWKV-Infer}  \cite{RWKV-Android}  \cite{rwkv-ncnn}  \cite{rwkv-qualcomm}  \cite{rwkv.jni}  \cite{RWKV6_Keras_Operator}  \cite{RWKV_Role_Playing_API}  \cite{nanoRWKV}] 
            ]
        ]
        [Applications of the \\ RWKV Model(Sec.~\ref{sec:applications}),text width=9em
          [Natural Language \\ Generation,text width=7em 
          [Chatbot~\cite{rwkv_chatbot,RWKV-wechat-bot,rwkv_chat_command_line, lommatzsch2023combining, eloise, Meow-AI, Espritchatbot-RASA-RWKV, RWKV-RAG, ChatRWKV-in-wechat-Version-1,Infofusion} \\Role-playing ~\cite{RWKV-Drama, RWKV_Role_Playing_with_GPT-SoVITS} /  Writing robot~\cite{AI-Writer} \\ Others\cite{notgpt} \cite{ykkz} \cite{Easy_RWKV_webui}  \cite{Espitchatbot-RASA-RAVEN}]
          ]
          [Natural Language \\ Understanding,text width=7em 
           [Virtual Assistant~\cite{lala_rwkv_chatbot_2.0} /  PDF Query Systems~\cite{pdf_parsing} \\ Machine Translation~\cite{liu2023approach} / OnlySports~\cite{chen2024onlysportslm} \\ Classification~\cite{RWKV-Classification} / Write novel~\cite{novel-rwkv_demo} \\  Others \cite{ai-town-rwkv-proxy, RavenWhisperer, wenda-RWKV, AVATARIO, rwkv_kg}]
          ]
          [Other NLP \\ tasks,text width=4em 
           [CLOPPS~\cite{xiacontrastive} / MSRWKV-2DTCN~\cite{hao2024multi} / Stock Price Prediction ~\cite{Stock-Prediction-Using-RWKV, dong2024dft, cao2024matcc} \\ RWKVCC~\cite{zhou2023code}  / Code Completion~\cite{zhou2023rwkv}   / Mod-RWKV~\cite{yildirim2024experimentation} / GoldFinch~\cite{goldstein2024goldfinch} \\ RWKV-TS~\cite{hou2024rwkv} / TIM~\cite{wang2024temporal}  / RRWKV~\cite{wang2023rrwkv}  / LKPNR~\cite{chen2024lkpnr}  \\ Decision-RWKV~\cite{dong2024optimizing} / ProSG~\cite{luo2023prosg} / SpikeGPT~\cite{zhu2023spikegpt}  \\
           Others~\cite{tsuruga2023general} \cite{dumitru2024enhancing} \cite{suzuki2024sensorimotor} \cite{drichel2024transfer}
           ] 
          ]
          [Computer \\ Vision,text width=4em 
           [BSBP-RWKV~\cite{zhou2024bsbp} / Restore-RWKV~\cite{yang2024restore} / RWKV-based MIL~\cite{ji2023rnn} / LION~\cite{liu2024lion} \\ PointRWKV~\cite{he2024pointrwkv}  / OccRWKV~\cite{wang2024occrwkv} / TLS-RWKV~\cite{zhu2024tls}  / RuleVM~\cite{jiang2024explicit}  \\ LineRWKV~\cite{valsesia2024hybrid} / RWKV-SAM~\cite{yuan2024mamba} / Vision-RWKV~\cite{duan2024vision} /  VisualRWKV~\cite{li2024visualrwkv} \\ LSTM CrossRWKV~\cite{yin2024video} / RWKV-CLIP~\cite{gu2024rwkv} / SDiT~\cite{yang2024sdit} / Social-CVAE~\cite{xu2023social} \\
           Diffusion-RWKV~\cite{fei2024diffusion} /  RWKV-IR~\cite{du2024exploring} / StyleRWKV~\cite{dai2024stylerwkv} \\
           Others~\cite{FaceRWKV} \cite{rwkv-denoise}] 
          ]
          [Audio and \\ Music Tasks,text width=5em 
           [Android RWKV MIDI~\cite{Android-RWKV-MIDI} / RWKV-v4 MIDI~\cite{RWKV-v4-MIDI} / RWKV-v4 music models ~\cite{Procedural-Purgatory} \\ Music-Diff~\cite{liu2024perturbing} / Voice Activity Detection~\cite{zuo2023advancing} / RWKV-Transducer~\cite{an2023exploring} \\ RWKV-ASR~\cite{RWKV-ASR} / AudioRWKV~\cite{AudioRWKV} / RWKV-MusicGenerator~\cite{RWKV-MusicGenerator} \\ Music-Genre-Classification-RWKV~\cite{Music-Genre-Classification-RWKV} \ RWKV-SpeechChat~\cite{RWKV-SpeechChat}] 
          ]
          [Web \\ Application,text width=5em 
           [MoZi-RWKV~\cite{MoZi-RWKV} / RWKV-3B-7B-Webui-GPT-SoVITS~\cite{RWKV_3B_7B_Webui_GPT-SoVITS} / AI00-Server~\cite{ai00_server}  \\ RWKV-v4-Web~\cite{rwkv-v4-web} / RWKV-Role-Playing~\cite{RWKV_Role_Playing}  / RWKV-QQBot-BackEnd~\cite{RWKV_QQBot_BackEnd}\\ Web-RWKV-Axum~\cite{web-rwkv-axum} / WebChatRWKVstic~\cite{WebChatRWKVstic}  / Chatux-Server-RWKV~\cite{chatux-server-rwkv} \\ ChatRWKV-Flask-Api~\cite{ChatRWKV-flask-api} / ChatRWKV-Webui~\cite{ChatRWKV-webui} / RWKV-Api~\cite{RWKV-api} \\ RWKV.cpp-Webui-Macos~\cite{rwkv.cpp_webui_Macos}  / RWKV-Role-Playing-UI~\cite{RWKV_Role_Playing_UI} \\ RWKV.cpp-Webui-GPT-SoVITS~\cite{rwkv.cpp_webui_GPT-SoVITS} ] 
          ]
          [Evaluation of \\ RWKV Models,text width=6em 
           [REGBENCH~\cite{akyurek2024context} / LongctxBench~\cite{yuan2024kv} / LooGLE~\cite{li2023loogle} / RULER~\cite{hsieh2024ruler} \\ S3EVAL~\cite{lei2023s3eval} / MAGNIFICO~\cite{patel2023magnifico}  / MANGO~\cite{ding2024mango} / SuperCLUE~\cite{xu2023superclue} \\ Zhujiu~\cite{zhang2023zhujiu} / BIPIA~\cite{yi2023benchmarking} / CoDI-Eval~\cite{chen2024benchmarking} / VasE~\cite{lanctot2023evaluating} / CMATH~\cite{wei2023cmath}  \\ Head-to-Tail~\cite{sun2023head}  / PRE~\cite{chu2024pre} / Freshbench~\cite{zhu2024evaluating} / LongICLBench~\cite{li2024long} \\ LongBench~\cite{LongBench_RWKV} / Long Range Arena~\cite{rwkv-long-range-arena} / MT-bench~\cite{MT_BENCH_RWKV}\\
           Others~\cite{leeattention} \cite{michaelov2024revenge} \cite{grmveliki} \cite{zhou2024benchmarking} \cite{paulo2024does} \cite{vacareanu2024words} \cite{huang2024well} \cite{compare_llms}
           ] 
          ]
          [Others,text width=3em 
           [Spiking Mixer~\cite{chenspiking} / MiniRWKV-4~\cite{MiniRWKV-4} / RWKV-RAG~\cite{RWKV-RAG} \\ VisualRWKV~\cite{hou2024visualrwkv}  / Dlip-RWKV~\cite{Dlip-RWKV} / RWKV-LM-EXT~\cite{RWKV_LM_EXT} \\  RWKV-Reward-Enhanced~\cite{rwkv-reward-enhanced} / RWKV-Twitter-Bot-Detection~\cite{Bot-Ani-RWKV-twitter-bot-detection}  \\ UnifiedImplicitAttnRepr~\cite{zimerman2024unified} / RWKV-PEFT~\cite{RWKV-PEFT} \cite{RWKV-PEFT-Simple} / RWKV5-InfctxLM~\cite{RWKV5-infctxLM} \\ DecisionRWKV~\cite{DecisionRWKV} / RWKV-Nonogram~\cite{RWKV-nonogram} / RWKV-RUL~\cite{RWKV_RUL}  \\ TrainChatGalRWKV~\cite{TrainChatGalRWKV} / TinyRWKV~\cite{tinyrwkv} / Simple-RWKV~\cite{simple_rwkv} \\ RWKV-Runner~\cite{RWKV-Runner} / Llama-node~\cite{llama-node} / Godot-rwkv~\cite{godot-rwkv} \\ RWKVRaven-Alpaca~\cite{HF-For-RWKVRaven-Alpaca}  / RWKV-Reward~\cite{rwkv_reward} / Decision-RWKV~\cite{Decision-RWKV} \\ RWKV-LM-RLHF~\cite{RWKV-LM-RLHF} / RWKV-LM-State~\cite{RWKV-LM-State-4bit-Orpo} / MARWKV~\cite{thapa2024modern} \\ RWKV-Finetune~\cite{RWKV-finetune-script} / RWKVWorld-LoraAlpaca~\cite{HF-For-RWKVWorld-LoraAlpaca}  \\ RWKV-LM-LoRA~\cite{RWKV-LM-LoRA-ja} / RWKV-Light-API~\cite{RWKV-Light-API} / RWKV-StateTuning~\cite{RWKV-StateTuning} \\ ChunkRWKV6~\cite{chunkRWKV6} / RWKV-LM-Interpretability~\cite{RWKV-LM-Interpretability-Research} \\ web-RWKV-Inspector~\cite{web-rwkv-inspector} / RWKV-Umap\cite{rwkv_umap}  / RWKV-Tensorrt\cite{rwkv-tensorrt} \\ LLMFarm~\cite{LLMFarm} / RWKV-Chains~\cite{RWKV_chains} / RWKV-Ouroboros~\cite{RWKV-Ouroboros-app} \\ AltaeraAI~\cite{AltaeraAI} / RWKVinLLAMA~\cite{RWKVinLLAMA} / GPTQ-for-RWKV~\cite{GPTQ-for-RWKV}  \\ RWKV-LM-AMD-Radeon-ROCm~\cite{RWKV-LM-AMD-Radeon-ROCm-hip} / InnerDNN~\cite{InnerDNN} / EasyChat-Server~\cite{EasyChat-Server} \\ ChatRWKV-PC~\cite{ChatRWKV_PC} / Generative Calibration~\cite{jiang2023generative} / RWKV- DPA~\cite{fu2024linear} \\ RWKV-edge~\cite{choe2024rwkv} / PCF-RWKV~\cite{li2024pcf} / L3TC~\cite{zhang2024l3tc} /PlantFlowerDatasets~\cite{PlantFlowerDatasets}] 
          ] 
        ]
        [Future Directions and \\ Challenges (Sec.~\ref{sec:challenges}),text width=9em
            [Expanding RWKV's Capabilities \\ in Future Applications,text width=12em
                [(1) Long sequence processing capability \\ (2) Multi-modal and Cross-modal Learning \\ (3) Parameter-efficient Finetuning]
            ]
            [Security and \\ Stability,text width=5em
                [(1) Safety\&Robustness \\ (2) Bias and fairness \\ (3) Privacy]
            ]
            [Optimizing RWKV for \\ Hardware Acceleration,text width=9em
            ]
        ]
    ]
\end{forest}

%% file: tex/background.tex
\section{Background}\label{sec:background}

\subsection{Recurrent Neural Networks (RNNs)}
Recurrent Neural Network (RNN) ~\cite{graves2012long} is a powerful tool for processing sequence data. It is capable of internally maintaining a state, which enables it to capture the dynamic features of time series data. The fundamental concept underlying RNN is to leverage the dependency among consecutive elements within a sequence and transmit information via cyclic connections. This allows the network to retain prior information and subsequently utilize the prior information for further computations, thereby endowing it with the ability to handle sequential data in a more effective and contextually aware manner.

The working principle of RNN can be described by the following mathematical model:

\begin{align}
    h_t = \tanh(W_{hh} \cdot h_{t-1} + W_{hx} \cdot x_t + b_h) \\
    y_t = W_{ho} \cdot h_t + b_o
\end{align}

Where $h_t$ represents the hidden layer state at time $t$, $x_t$ represents the input at time $t$, and $y_t$ represents the output at time $t$. These two equations demonstrate the process of merging the current input with the preceding hidden state to derive the present hidden state. Subsequently, they also illustrate how this hidden state is transformed into an output.

While RNN models are effective at handling time series data, they struggle with recognizing long-term dependencies. This issue arises primarily due to the phenomena of vanishing or exploding gradients. In RNN models, the issue of gradient disappearance arises when gradients are multiplied by the weight matrix across time steps. When the eigenvalue of this weight matrix is less than 1, it results in the exponential diminishment of gradient values over time, leading to vanishing gradients. Conversely, when the eigenvalue exceeds 1, the gradient values increase exponentially with time, resulting in gradient explosion.

The structure of RNN models imposes a restriction on their capability to learn long-term dependencies. In typical RNN models, each time step's hidden state is influenced solely by the hidden state from the preceding time step. Consequently, RNN models struggle to effectively capture and understand dependencies that span across extended time periods.

In order to solve the problem of gradient disappearance or gradient explosion that may occur when RNN processes long sequences, researchers have proposed several variants of RNN:
Long short-term memory network (LSTM)~\cite{hochreiter1997long}: By introducing a gating mechanism to control the flow of information, it can capture long-term dependencies.
Gated recurrent unit (GRU)~\cite{chung2014empirical}: The structure is simpler than LSTM, but it can provide similar performance.
Bidirectional recurrent neural network (Bi-RNN)~\cite{schuster1997bidirectional}: By processing the forward and reverse information of the sequence at the same time, it can more comprehensively understand the patterns in the sequence.

Recently, with increasing attention on overcoming the scalability limitations of Transformer-based architectures, the potential of RNN-based variants has garnered renewed interest. Models such as minLSTMs and minGRUs~\cite{feng2024rnnsneeded}, xLSTM~\cite{beck2024xlstm}, and Aaren~\cite{feng2024rnnsneeded} show that addressing the limitations of traditional RNNs can make these models perform as well as or even better than Transformers in various tasks. These developments highlight the enduring significance and promising potential of RNN-based approaches in advancing modern deep learning.

\subsection{Transformers}
In RNNs, the computation of each time step is based on the output of the previous step, which requires sequential processing and prohibits parallel computation. Conversely, the Transformer architecture employs the Self Attention mechanism to process all time steps simultaneously, enabling parallel computation. Unlike conventional recurrent neural network models like LSTM and GRU, the Transformer~\cite{vaswani2017attention} eschews the recursive architecture entirely. Instead, it leverages self-attention to establish dependencies across all positions within the input sequence, thereby enhancing both training speed and efficiency. 

Attention mechanisms have the capability to identify and record the relationships across all input and output tokens. In Self-Attention, the input is denoted by the matrix $x$. This undergoes a linear transformation using the matrices $W_q$, $W_k$, and $W_v$, allowing the calculation of the Query $Q$, Key $K$, and Value $V$. It is important to highlight that each row within the matrices $x$, $Q$, $K$, and $V$ corresponds to an individual word.
\begin{align}
    Q = x \cdot W_q,  K = x \cdot W_k,  V = x \cdot W_v 
\end{align}
where $W_q$, $W_k$, $W_v$ are the trainable parameters.
After obtaining the matrices $Q$, $K$, $V$, it can calculate the output of Self-Attention. The calculation formula is shown as follows:
\begin{align}
    Attn(Q, K, V) = softmax(\frac{QK^T}{\sqrt{d_k}})V
\end{align}
where $d_k$ is the dimension of the key vectors. It scales the result by $\sqrt{d_k}$
in order to prevent the inner product from being too large.

In a single Attention mechanism, a singular representation space is derived. When utilizing multiple attentions, it is possible to derive several distinct representation spaces. This concept leads to the development of Multi-Head Attention by integrating several Self-Attentions. Each Attention employs unique weight matrices for Query, Key, and Value, with these matrices initialized in a random manner. Training subsequently projects word embeddings into diverse representation spaces. The mathematical expression representing the multi-head attention mechanism is as follows:
\begin{align}
    &Q_i=x\cdot W_{q_i},K_i=x\cdot W_{k_i},V_i=x\cdot W_{v_i}, \\
    &head_i=Attention(Q_i,K_i,V_i), i\in[1,n], \\
    &\scalebox{0.9}{$
    MultiHead(Q,K,V)=Concat(head_1,\ldots,head_n)W_o
    $}
\end{align}
where $n$ means the number of attention heads, $Concat$ means concatenate and $W_o$ is the projection matrix which is a trainable parameter.
Multi-head attention enables the simultaneous processing of input sequences from various perspectives, enhancing the model's capacity to comprehend and identify intricate dependencies. This capability substantially boosts the Transformer model's effectiveness in handling natural language processing tasks as well as other types of sequence data processing.

\subsection{Attention Free Transformer (AFT)}
The attention mechanism in Transformers excels at capturing long-term dependencies and emphasizing pertinent elements within the input sequence. Nevertheless, the self-attention mechanism demands matrix multiplication involving the transpose of the query and key. This operation exhibits quadratic space complexity relative to sequence length, rendering it inefficient for handling extended sequences effectively.
Attention Free Transformer (AFT)~\cite{zhai2021attention}, represents a Transformer variant that does not rely on matrix multiplication. This change leads to a reduction in memory complexity—from $O(T^2d+Td)$ to $O(Td)$, as noted by \cite{peng2023rwkv}—and a decrease in time complexity—from $O(T^2d)$ to $O(Tsd)$, according to \cite{peng2023rwkv}. Here, $T$ denotes the sequence length, $d$ stands for the embedding dimension, and $s$ represents the size of a local window specific to AFT. Two primary AFT variants exist: AFT-full and AFT-local. The latter, AFT-local, arose from the author's observation that attention typically exhibits pronounced local patterns; thus, utilizing a reduced attention window was deemed advantageous.

Same as Transformer, for the input feature $x$, AFT-full uses three different linear layers to calculate Query $Q$,  Key $K$, and Value $V$, 
$Q\in \mathbb{R}^{T\times d}$, $K\in \mathbb{R}^{T\times d}$, $V\in \mathbb{R}^{T\times d}$, where $T$ is the maximum length of the sequence,
$d$ is the number of hidden layer nodes. Unlike Transformer, AFT adds a learnable positional encoding $w_{t^{\prime},t}$ to $K_{t^{\prime}}$, where $1 < t^{\prime}< T$. And use softmax to normalize $K$ with position encoding.
\begin{align}
    \mathrm{Weighted}(\mathbf{K}_{t^{\prime}})=\frac{\exp(\mathbf{K}_{t^{\prime}}+w_{t,t^{\prime}})}{\sum_{t^{\prime}=1}^T\exp(\mathbf{K}_{t^{\prime}}+w_{t,t^{\prime}})}
\end{align}

Subsequently, the sigmoid activation function is applied to normalize $Q_t$, resulting in the weight matrix $\alpha_t^{\prime}$, which is derived by conducting a dot multiplication with $\mathrm{Weighted}(\mathbf{K}_{{t^{\prime}}})$:
\begin{align}
\alpha_t^{\prime} &=\sigma_Q(Q_t)\odot\text{Weighted}(\mathbf{K}_{t^{\prime}}) \\
&=\sigma_Q(Q_t)\odot\frac{\exp(\mathbf{K}_{t^{\prime}}+w_{t,t^{\prime}})}{\sum_{t^{\prime}=1}^T\exp(\mathbf{K}_{t^{\prime}}+w_{t,t^{\prime}})}
\end{align}

Finally, the AFT-full's attention mechanisms $Attn(Q,K, V)_t=\mathbf{Y}_{t}$ is calculated as follow:
\begin{align}
\mathbf{Y}_{t}& =\sum_{t^{\prime}=1}^T\alpha_{t^{\prime}}\mathbf{V}_{t^{\prime}} \\
&=\sum_{t^{\prime}=1}^T\sigma_Q(\mathbf{Q}_t)\odot\frac{\exp(\mathbf{K}_{t^{\prime}}+w_{t,t^{\prime}})}{\sum_{t^{\prime}=1}^T\exp(\mathbf{K}_{t^{\prime}}+w_{t,t^{\prime}})}\odot\mathbf{V}_{t^{\prime}} \\
&=\sigma_Q(Q_t)\odot\frac{\sum_{t^{\prime}=1}^T\exp(\mathbf{K}_{t^{\prime}}+w_{t,t^{\prime}})\odot V_{t^{\prime}}}{\sum_{t^{\prime}=1}^T\exp(\mathbf{K}_{t^{\prime}}+w_{t,t^{\prime}})}\label{aft-full}
\end{align}

Analyzing equation \ref{aft-full} along with the  implementation of AFT-full reveals that the complexity of AFT-full is $O(T^2d)$, similar to that of the Transformer. However, the distinct feature of AFT-full is that this complexity arises from the incorporation of $w$ into the computations, which allows for modifying the style of $w$ in order to enhance AFT's speed. The authors of AFT introduced AFT-local as a refinement. In this variant, any position embedding values that extend beyond the window $s$, with the condition $s\leq T$, are assigned a value of zero.
\begin{align}
    \left.w_{t,t^{\prime}}=\left\{\begin{array}{ll}w_{t,t^{\prime}},&\mathrm{if~}|t-t^{\prime}|<s\\0,&\mathrm{otherwise}\end{array}\right.\right.
\end{align}
AFT-local assigns a value of 0 to the position embedding outside the window. However, since these zeroed values remain part of the entire computational process, there is no reduction in the computational load. Thus, the time complexity for AFT-local remains $O(Tsd)$.

%% file: tex/rwkv.tex
\section{RWKV}\label{sec:rwkv}

The presence of a cyclical architecture in RNNs~\cite{graves2012long} is responsible for their elevated computational complexity and reduced training efficiency. This architecture also makes them susceptible to issues such as gradient vanishing and gradient exploding, thereby limiting the scalability of RNNs. In contrast, the Transformer~\cite{vaswani2017attention} employs a self-attention mechanism, enabling it to handle entire sentences at once. This allows for improved training efficiency and makes parallel processing possible. Nonetheless, Transformers are characterized by high computational complexity, with their time complexity being $O(T^2d)$, making them less suitable for handling long sequences efficiently.
Grounded in the classic Transformer framework, AFT~\cite{zhai2021attention} replaces matrix multiplication with a technique akin to Residual Connection. This change allows AFT to bypass the attention mechanism, thereby lessening the space requirements. Consequently, AFT is capable of computing the attention weight with substantially reduced space complexity compared to the Transformer model; however, it does not achieve major improvements in reducing time complexity.

RWKV~\cite{peng2023rwkv} was developed in response to the challenges posed by RNNs, Transformers, and AFT issues, primarily focusing on enhancing the linear attention mechanism. This innovation aims to address the difficulty of parallelizing RNNs. The model maintains a time complexity comparable to that of RNNs while achieving performance effects akin to Transformers.

Currently, the official versions of RWKV mainly include three types: RWKV-4, RWKV-5 (Eagle), and RWKV-6 (Finch). Among them, the commonly used version is RWKV-4. In addition, the latest RWKV-7 (Goose) has released a preview version and the corresponding code, but the specific document has not yet been made public yet\footnote{\url{https://www.rwkv.com}} . 

\subsection{RWKV-4}
\begin{figure}[htbp]
\centering
\includegraphics[width=\linewidth]{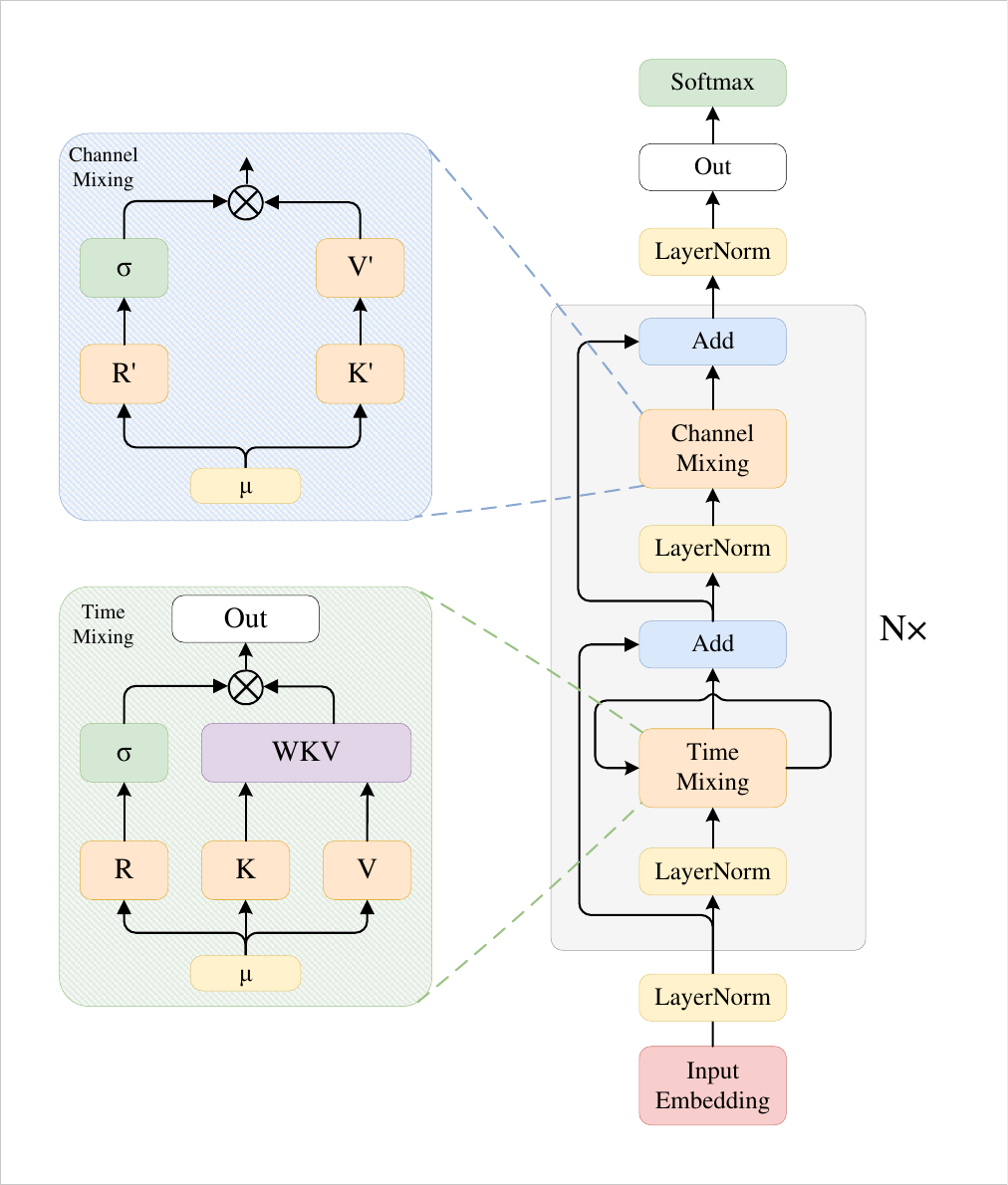}
\caption{The structure of the RWKV model consists of stacked residual blocks, where each block is made up of a time-mixing sub-block and a channel-mixing sub-block, incorporating recurrent elements to capture past information.}
\label{fig:rwkv structure}
\end{figure}

RWKV-4~\cite{peng2023rwkv} represents the first version officially made available to the public, with RWKV-1, RWKV-2, and RWKV-3 existing as experimental predecessors. Referring to figure~\ref{fig:rwkv structure}, the architecture of the RWKV-4 model includes a series of stacked residual blocks. Each of these blocks contains sub-blocks dedicated to time-mixing and channel-mixing, which efficiently leverage previous information through an incorporated recursive framework.

The time-mixing block is designed for global interaction, akin to how self-attention functions in a Transformer. Initially, the term "RWKV" is composed of the following four letters:
\begin{itemize}
    \item $R$: Receptance vector, responsible for gauging the amount of past information that is permitted;
    \item $W$: Weight vector, includes a position weight decay vector and contains the model's trainable parameters;
    \item $K$: Key vector, analogous to the $K$ in the traditional attention mechanism;
    \item $V$: Value vector, analogous to the $V$ present in the traditional attention mechanism.
\end{itemize}
For time $t$, given word $x_t$ and the previous word $x_{t-1}$, the Time-Mix module formulas are as follows:
\begin{align}
r_{t} &=W_r\cdot(\mu_r\odot x_t+(1-\mu_r)\odot x_{t-1}), \label{r}\\
k_{t} &=W_k\cdot(\mu_k\odot x_t+(1-\mu_k)\odot x_{t-1}), \label{k}\\
v_{t} &=W_v\cdot(\mu_v\odot x_t+(1-\mu_v)\odot x_{t-1}), \label{v}\\
wkv_t &=\scalebox{1}{$\frac{\sum_{i=1}^{t-1}exp(-(t-1-i)w+k_i)\odot v_i+exp(u+k_t)\odot v_t}{\sum_{i=1}^{t-1}exp(-(t-1-i)w+k_i)+exp(u+k_t)}$}\label{wkv} \\ 
o_t &=W_o\cdot(\sigma(r_t)\odot wkv_t) \label{rwkv}
\end{align}
The terms $r_{t}$, $k_{t}$, and $v_{t}$ utilized in this context are analogous to the $Q$, $K$, and $V$ components found in AFT or Transformer architectures. A notable distinction in the computation of $R$, $K$, and $V$ as compared to the Transformer model is that the input $x$ is not merely the embedding of the current token. Instead, it represents the weighted sum of the current token's embedding and that of the preceding token. Additionally, the computation of $wkv_t$ is linked to the attention mechanism's implementation. This implementation represents a linear interpolation utilizing both historical and present moment data, and it is important to note its exponential nature. The relationship between the current token and all preceding tokens is characterized by an exponential decay sum. This specific aspect endows $wkv_t$ with the features of linear attention. Notably, there exist two primary distinctions between equation \ref{wkv} and equation \ref{rwkv} as well as AFT's equation \ref{aft-full}:

1. AFT modifies the bias $w_{t^{\prime},t}$ from being based on absolute position to being based on relative position, resulting in the need to train only a single parameter $w$ vector.

2. The parameter $u$ is introduced to handle the current position independently within the process.

The RWKV model employs Time Mixing to capture the relationship between tokens, and the Channel Mixing module models RWKV by exploring the dimensions of the hidden layer that correspond to individual tokens.
The channel-mixing block utilizes equations \ref{r} and \ref{k} again to determine a fresh set of $r_t'$ and $k_t'$ using the result from the time-mixing block. Subsequently, the final output is computed in the following manner:
\begin{align}
    r_t'&=W_r'\cdot(\mu_r'\odot x_t+(1-\mu_r')\odot x_{t-1}),\\
    k_t'&=W_k'\cdot(\mu_k'\odot x_t+(1-\mu_k')\odot x_{t-1}), \\  o_t'&=\sigma(r_t')\odot(W_v'\cdot\max(k_t',0)^2).
\end{align}

Channel Mixing conducts integration within the feature dimension. Given that the feature vector has a dimension of $d$, each element in these dimensions must gather data from other dimensions to refresh its own value. In this context, each feature vector dimension is referred to as a ''channel''.

The RWKV architecture combines the strengths of both Transformers and RNNs. Unlike conventional RNNs, it offers the benefit of having a stable gradient conducive to deeper architecture, similar to that of Transformers. Additionally, RWKV excels in reasoning efficiency. RWKV redefines the attention framework by adopting linear attention, substituting the usual dot-product interaction between tokens with a more efficient attention directed at channels. This adaptation results in a computational complexity of $O(Td)$ and decreases the memory complexity down to $O(d)$.

\subsection{RWKV-5 (Eagle)}
The sequential models Eagle (RWKV-5) and Finch (RWKV-6)~\cite{peng2024eagle} have been developed as advancements of the RWKV (RWKV-4) architecture. In these models, enhancements have been incorporated into the architecture design, such as multi-headed matrix-valued states and dynamic recursive mechanisms, which bolster the models' expressive capabilities while preserving the inference efficiency typical of RNNs. Compared with RWKV-4, RWKV-5 and RWKV-6 mainly made changes to the time mixing block.

Eagle (RWKV-5)~\cite{peng2024eagle} enhances the architecture and learning decay strategy derived from RWKV-4~\cite{peng2023rwkv} through the adoption of expressive multi-head matrix-valued states rather than the traditional vector-valued states. It also includes a reconfiguration of receptive states and incorporates supplementary gating mechanisms.

Eagle introduced a new notation $lerp$, which means the linear interpolation between $x_t$ and $x_{t-1}$ used in RWKV-4 and
it is defined as:
\begin{align}
    \operatorname{lerp}_\square(a,b)=a+(b-a)\odot\mu_\square 
\end{align}
where each $\mu_\square\in\mathbb{R}^D$ is a learnable vector.
The formula of Eagle Time Mixing can be written as follows:
\begin{align}
    &\square_t=\operatorname{lerp}_{\square}(x_t,x_{t-1})\boldsymbol{W}_{\square},\quad\square\in\{r,k,v,g\}\\
    &w=\exp(-\exp(\omega))\\
    &\boldsymbol{w}\boldsymbol{k}\boldsymbol{v}_t = \operatorname{diag}(u) \cdot k_t^\mathrm{T} \cdot v_t \\
    &\quad \quad \quad + \sum_{i=1}^{t-1} \operatorname{diag}(w)^{t-1-i} \cdot k_i^\mathrm{T} \cdot v_i \in \mathbb{R}^{(D/h)\times(D/h)}, \nonumber \\
    &\scalebox{0.89}{$o_t=\operatorname{concat}\big(\operatorname{SiLU}(g_t)\odot\operatorname{LayerNorm}(r_t\cdot\boldsymbol{w}k\boldsymbol{v}_t)\big)\boldsymbol{W}_o\in\mathbb{R}^D$}
\end{align}
LayerNorm is applied individually to each of the $h$ heads, which can be seen as analogous to the Group-Norm applied over $h$ groups. Additionally, it should be highlighted that $w$ is derived using the formula $w=\exp(-\exp(\omega))$, with $\omega\in\mathbb{R}^{D/h}$ representing the trainable parameters for each head. This computation ensures that $w$ remains within the range $(0,1)$, thus assuring that $\operatorname{diag(w)}$ functions as a contraction matrix. 
Then, The $\boldsymbol{w}\boldsymbol{k}\boldsymbol{v}_t$ attention calculation can alternatively be written in a recurrent form:
\begin{align}
 \boldsymbol{wkv}^{\prime}=s+\mathrm{diag}(u)\cdot k^\mathrm{T}\cdot\nu\\
    s^{\prime}=\mathrm{diag}(u)\cdot s+k^\mathrm{T}\cdot\nu 
\end{align}

\subsection{RWKV-6 (Finch)}
Finch (RWKV-6)~\cite{peng2024eagle} enhances the architecture's expressiveness and adaptability by incorporating innovative data-driven functions into the time mixing and token shifting modules, such as parameterized linear interpolation. Furthermore, Finch (RWKV-6) introduces a novel application of low-rank adaptation functions~\cite{hu2022lora}, allowing the weight matrices to be trainable and effectively augment the learned data decay vectors in a way that is sensitive to the context.

The token shift of Finch is $ddlerp$, which means data dependent linear interpolation between $x_t$ and $x_{t-1}$, and it is defined as:
\begin{align}
&\text{lora}_{\square}(x) =\lambda_\square+\tanh(xA_\square)B_\square  \\
&\scalebox{0.95}{$\mathrm{ddlerp}_{\square}(a,b) =a+(b-a)\odot\text{lora}_\square(a+(b-a)\odot\mu_x) $}
\end{align}
In this setup, both $\mu_x$ and each $\lambda_\square$ bring in a trainable vector with dimension $D$, while each $A_\square\in\mathbb{R}^{D\times32}$ and $B_\square\in\mathbb{R}^{32\times D}$ represent new trainable weight matrices. The LoRA mechanisms~\cite{hu2022lora} described above enable the cost-effective enhancement of learned vectors, similar to those in Eagle, by incorporating additional offsets that are defined by the incoming input. By integrating this innovative form of Token Shift with data-dependence, Finch aims to extend the model's capabilities beyond the Token Shift method utilized in RWKV-4/Eagle. The model now computes the distribution of new and existing data per channel based on the input at both the present and preceding time steps. Thus, Finch's time mixing block is defined accordingly:
\begin{align}
&\square_t =\mathrm{ddlerp}_\square(x_t,x_{t-1})W_\square,\quad\square\in\{r,k,\nu,g\} \\
&d_{t} =\mathrm{\text{lora}}_d(\mathrm{ddlerp}_d(x_t,x_{t-1})) \\
&w_t =\exp(-\exp(d_t)) \\
&\boldsymbol{wkv}_t =\mathrm{diag}(u)\cdot k_t^\mathrm{T}\cdot\nu_t     \\ 
&\quad \quad \quad +\sum_{i=1}^{t-1}\mathrm{diag}\left(\sum_{j=i+1}^{t-1}w_j\right)\cdot k_i^\mathrm{T}\cdot\nu_i\in\mathbb{R}^{(D/h)\times(D/h)} \nonumber\\
&\scalebox{0.89}{$o_{t} =\operatorname{concat}\left(\operatorname{SiLU}(g_t)\odot\operatorname{LayerNorm}(r_t\cdot\boldsymbol{wkv}_t)\right) \boldsymbol{W}_o\in\mathbb{R}^D $}
\end{align}
Unlike in Eagle, $w_t$ of Finch is not static across the sequence.
The fundamental transformation in Finch involves a shift to decay, allowing each component of $w_t$ to independently fluctuate over time in accordance with data, unlike the prior model which employed a static learned vector. The repetitive structure of the $\boldsymbol{wkv}_t$ attention computation remains identical to that of Eagle.

\subsection{Compare RWKV with Other improved models based on Transformer}
Linear Transformers~\cite{wang2020linformer} represent a variant of the Transformer architecture. While the conventional Transformer relies on the Self-Attention mechanism, Linear Transformers focus on approximating or enhancing this mechanism through a more efficient linear approach. Their primary goal is to decrease computational complexity and boost the model's training and inference efficiency without sacrificing the transformative sequence processing capabilities inherent to Transformer models. Linear Transformers employ kernel methods to approximate self-attention. For instance, in linear self-attention, they utilize a low-rank matrix to approximate the query-key ($Q$-$K$) product found in standard self-attention mechanisms.
Consider $Q$ as the query matrix and $K$ as the key matrix. In standard self-attention, the expression $QK^{T}$ involves a quadratic computation. Linear transformers can incorporate a low-rank matrix $M$, altering the operation to $QMK^{T}$, where $M$ has a lower rank. Selecting a suitable $M$ allows the computational complexity of this operation to decrease to a linear level. Utilizing kernel methods, linear transformers achieve a computational complexity of \(O(Td^{2})\). In contrast, the computational complexity of RWKV is \(O(Td)\), indicating that linear transformers require more computation. Additionally, as linear transformers approximate the typical self-attention mechanism, there could be a diminution in the model's representational capabilities.

In addition to RWKV, Mamba ~\cite{mamba} is another frequently encountered and notable improved model developed from the Transformer architecture.
Mamba leverages state-space models (SSMs), setting itself apart by effectively incorporating Structured State Space (S4) models into a large language model (LLM) framework. This inventive approach allows Mamba to achieve linear scalability concerning sequence length in complexity, marking a significant advancement compared to the quadratic complexity typically associated with traditional Transformer-based models. The carefully crafted architecture of Mamba, which incorporates selective SSM layers, enhances both computational efficiency and flexibility. Compared to RWKV, Mamba achieves better results in sequence segmentation. However, it demands greater computational resources. Conversely, RWKV excels by offering greater efficiency, quicker inference times, and reduced memory usage during operation~\cite{yuan2024mamba}.

Retentive Network (RetNet)  ~\cite{sun2023retentive} can simultaneously meet the three requirements of parallel training, low-cost inference, and good performance.
By introducing the Retention mechanism, RetNet effectively solves the problems of computational complexity and memory occupation that Transformer has when dealing with long sequence data. Compared with the Attention mechanism, the Retention mechanism is more efficient during the calculation process and can significantly reduce the consumption of computational resources and memory.
When processing sequence data, RetNet combines the parallel representation and the recurrent representation to achieve a balance between efficiency and performance.
In addition, RetNet can generate multi-scale sequence representations. This means that it can understand sequence data from different granularities or levels.
As a work emerging after RWKV, compared with RWKV, RetNet's retention mechanism preserves high-dimensional states to encode sequence information, thereby further enhancing the expressive ability and leading to better performance.

Hyena~\cite{poli2023hyena} is a subquadratic
drop-in replacement for attention constructed by interleaving implicitly parametrized long convolutions
and data-controlled gating. Unlike traditional RNNs and some Transformer-based architectures, the Hyena Hierarchy processes data through a hierarchical structure. The Hyena hierarchy consists of two operators (long convolution and element-wise multiplication gating) that are recursively defined by efficient quadratic primitives. Among them, the long convolution can effectively extract local features in sequences and, through the sliding window method, can capture similar patterns at different positions. The element-wise multiplication gating is generally located after the long convolution or between different layers in the Hyena hierarchy. Its main function is to screen and adjust the features extracted by the long convolution or the features transferred between different layers. The time complexity of Hyena is $O(ndt(\log t + d))$, and the space complexity is $O(td)$, where $n$ is the number of projections, $d$ is the dimension, and $t$ is the sequence length. In the case of long sequences, the growth of its computational complexity is relatively slow, enabling it to handle longer sequence data without a computational explosion.
As for RWKV, both the time complexity and the space complexity are $O(td)$. Compared with the traditional Transformer architecture, the computational efficiency has been significantly improved. Especially in the long-sequence inference scenario, its linear time complexity allows it to handle large-scale sequence data more efficiently. 

\subsection{Various Implementations}
The RWKV authors offer an official PyTorch implementation as detailed in version~\cite{PENG_RWKV-LM_2021}, encompassing the implementation specifics for RWKV-1 through RWKV-6. Their efforts are further extended with the ongoing development of a laboratory version of RWKV-7 (Goose). Additionally, they introduced a Chatbot framework called ChatRWKV~\cite{ChatRWKV}, designed to enable users to construct their own RWKV inference engines.

Beyond the primary implementation of PyTorch, RWKV has motivated a host of third-party versions developed in an array of programming languages and frameworks, thus showcasing its adaptability and wide-ranging utility. These implementations cover lightweight options in languages such as C~\cite{rwkv.cpp, rwkv.c}, C++~\cite{rwkv-cpp-accelerated, rwkv-cpp-server, rwkv_tokenizer_cpp}, and Fortran~\cite{rwkv.f90}. Additionally, RWKV has been integrated into modern programming languages including Julia~\cite{rwkv.jl} and Zig~\cite{rwkv.zig}. Notably, the ecosystems for Go~\cite{rwkv, rwkv_go, go-rwkv.cpp, verbaflow, rwkvalgorithm, rwkv-tokenizer-go} and Rust~\cite{web-rwkv, rwkvk-rs, rwkv-tokenizer, smolrsrwkv, rwkv-rust} exhibit strong support, offering models, tokenizers, and tailored platform optimizations.  Moreover, RWKV has been integrated into popular deep learning frameworks such as TensorFlow~\cite{rwkv-tensorflow, RWKV5-Tensorflow2.0, RWKV-Tensorflow2.0, Tensor-RWKV, rwkv-cuda, tensorflow-rwkv, RWKV-tf2}, Keras ~\cite{RWKV6-Keras, keras-rwkv}, and PaddlePaddle~\cite{rwkv-paddle, RWKV-v2-RNN-paddle}, allowing advanced features such as multi-GPU training and inference.  Additional contributions in JAX~\cite{tpu_rwkv, RWKV-LM-jax, rwkv-jax}, Node.js~\cite{RWKV-cpp-node, RWKV-tokenizer-node}, and MLX~\cite{mlx-rwkv} further highlight the model's versatility, establishing RWKV as a prominent tool for various computational and application needs.

RWKV's capabilities have been notably improved in various areas, including training, inference, and applications, through additional implementations. In the realm of training, time-parallel methods have been implemented, with more sophisticated options incorporating LoRA, 4-bit quantization, as well as support for platforms like CUDA and ROCm~\cite{RWKV-infctx-trainer, RWKV5-LM-LoRA, RWKV-infctx-trainer-LoRA}. For inference purposes, enhancements include Flash-Linear-Attention in RWKV v6~\cite{RWKV-Infer}, execution support on Android CPUs using ONNX~\cite{RWKV-Android}, compatibility with the NCNN framework~\cite{rwkv-ncnn}, and deployment on Qualcomm HTP with the QNN SDK~\cite{rwkv-qualcomm}. Moreover, a Java/JNI wrapper for RWKV.cpp has been developed to enable CPU-based inference with quantization, eliminating the need for Python~\cite{rwkv.jni}. Additionally, a kernel operator has been integrated into the bert4keras3 library to improve compatibility with RWKV6 models~\cite{RWKV6_Keras_Operator}. Implementations focused on specific applications include an API using the Flask framework, specifically tailored for role-playing scenarios~\cite{RWKV_Role_Playing_API}. Furthermore, inspired by NanoGPT, there exists an implementation utilizing RWKV's RNN architecture to deliver LLM performance on par with GPT~\cite{nanoRWKV}.

%% file: tex/applications.tex
\section{Applications of the RWKV Model}\label{sec:applications}
\begin{figure*}[htbp]
\centering
\includegraphics[width=\linewidth]{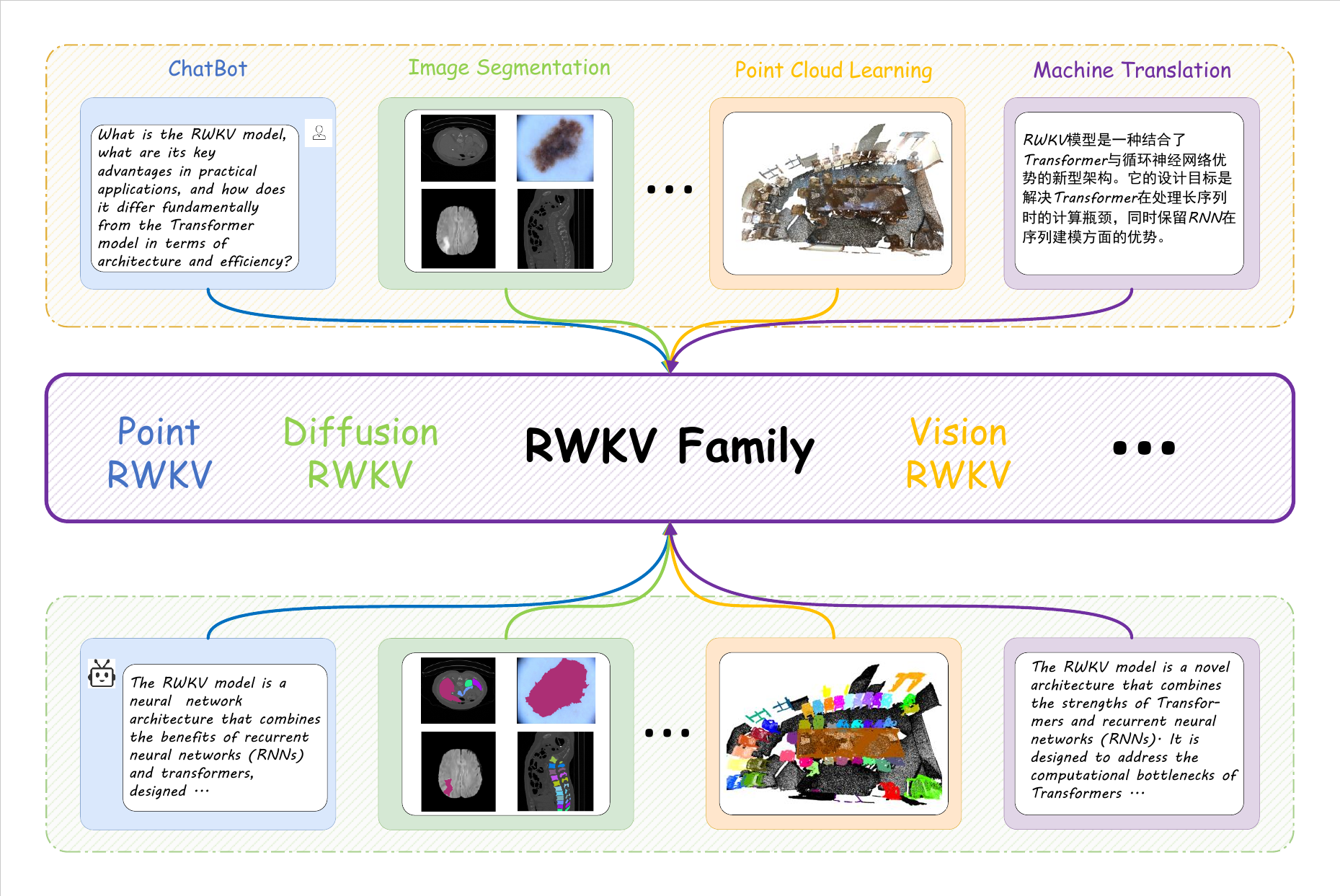}
\caption{Examples of downstream tasks utilizing RWKV-based models.}
\label{fig:application}
\end{figure*}

The RWKV model integrates the advantages of recurrent mechanisms with those of attention-based approaches. This unique architectural design enhances its efficiency and scalability, enabling it to adeptly tackle a wide range of problems. This chapter provides an overview of the practical applications of the RWKV model, focusing on its deployment in areas like natural language generation, natural language understanding, computer vision, audio and music tasks, and web-related tasks. Figure~\ref{fig:application} illustrates examples of downstream tasks that make use of RWKV-based models. Additionally, the discussion extends to the use of RWKV in other NLP tasks, where its performance is evaluated across diverse assessment frameworks.

\subsection{Natural Language Generation}
Natural Language Generation holds significant importance as an application for RWKV models, showcasing their ability to generate text that mirrors human-like communication. By leveraging the architecture of the RWKV model, it adeptly captures contextual subtleties and generates coherent narratives across diverse formats. The model excels at preserving contextual relevance while producing text that is fluid and captivating. Additionally, RWKV's adaptability enables it to adjust to various tones and styles, rendering it especially beneficial for tasks such as automated storytelling, content creation, and engaging chatbots.

Several investigations underscore the wide range of uses for RWKV in the field of natural language generation. As an example, RWKV has been effectively utilized in producing web novels spanning numerous genres, such as fantasy and romance~\cite{AI-Writer}. Additionally, the model has been integrated into multiple chatbot projects~\cite{rwkv_chatbot,RWKV-wechat-bot,rwkv_chat_command_line, lommatzsch2023combining}, such as a QQ chatbot~\cite{eloise}, a lightweight local chat AI~\cite{Meow-AI}, and a chatbot designed to respond to inquiries regarding Esprit University~\cite{Espritchatbot-RASA-RWKV}. In particular, a robust RWKV-based RAG deployment system facilitates the efficient management of local knowledge bases, while also providing a question-answering bot~\cite{RWKV-RAG}.

Within interactive settings, RWKV is used for role-playing scenarios~\cite{RWKV-Drama, RWKV_Role_Playing_with_GPT-SoVITS}, and also for generating WeChat responses via projects like Chatrwkv~\cite{ChatRWKV-in-wechat-Version-1, Infofusion}. Furthermore, RWKV is implemented in a Telegram LLM bot written in Rust, which can operate using either RWKV or Llama2 models~\cite{notgpt}. Chatbots with RWKV capability also facilitate multimedia interactions, such as sending voice messages and stickers~\cite{ykkz, Easy_RWKV_webui}. Additionally, chatbots using the Rasa NLU framework alongside RWKV-4-Raven can address frequently asked questions in French, English, and Tunisian Arabic~\cite{Espitchatbot-RASA-RAVEN}. Across these diverse applications, RWKV models manifest substantial potential in advancing natural language generation and refining human-computer interaction.

\subsection{Natural Language Understanding}
Understanding human language is a crucial capability of RWKV models, highlighting their proficiency in interpreting and analyzing it. By skillfully handling the intricate details of syntax, semantics, and context, RWKV models are particularly effective in performing tasks like machine translation, categorizing text, and providing answers to interactive questions.

RWKV models have shown encouraging outcomes in the field of machine translation. For instance, a translation algorithm based on RWKV for Mongolian-Chinese utilizes contrastive learning to address issues of semantic distortion and training instability that are often encountered in conventional approaches ~\cite{liu2023approach}. Furthermore, a compact language model related to sports, developed using RWKV-v6, has realized notable gains in accuracy, reaching levels of performance similar to those of larger models such as SomlLM and Qwen~\cite{chen2024onlysportslm}.

RWKV has been utilized in traditional NLU tasks, such as text classification~\cite{RWKV-Classification}, the creation of interactive narratives~\cite{novel-rwkv_demo}, and functioning as a virtual assistant on platforms including Discord~\cite{lala_rwkv_chatbot_2.0}. In addition, RWKV's application in gaming contexts has contributed to enhancing character dialogues, thus improving player interaction and immersion~\cite{ai-town-rwkv-proxy}. Moreover, it has been employed in interactive PDF query systems, where the RWKV model, alongside tools like Langchain and Streamlit, is used for tasks like summarization and information extraction~\cite{pdf_parsing}.

RWKV models have been successfully integrated into real-time applications, such as systems that listen for queries and interactively provide answers from the model~\cite{RavenWhisperer}, as well as question-and-answer systems that demonstrate its effectiveness in processing and interpreting complex language tasks~\cite{wenda-RWKV}. Furthermore, educational mobile applications aimed at engaging children through customizable avatars~\cite{AVATARIO} leverage RWKV technology. Another notable application involves using pure RWKV in knowledge graph implementations~\cite{rwkv_kg}, which enhances information retrieval and contextual understanding.

\subsection{Other NLP tasks}
The RWKV model has shown remarkable efficiency in a wide range of NLP tasks, such as forecasting, code generation, and content regulation. Its ability to adapt and deliver robust results across these varied applications underscores its transformative potential within the field of natural language processing.

Within the domain of predictive tasks, the RWKV model demonstrates considerable effectiveness. The CLOPPS (Clinical Outcomes Prediction with Partially Observed Data) model enhances classification capabilities by addressing the complexities linked to actual clinical datasets~\cite{xiacontrastive}. Similarly, a project drawing on data from the American Community Survey has effectively generated population forecasts for numerous towns across the United States ~\cite{tsuruga2023general}. In forecasting photovoltaic power generation, the MSRWKV-2DTCN model employs a multi-scale temporal convolutional network to adeptly capture both periodic and historical dependencies, thereby improving short-term prediction accuracy~\cite{hao2024multi}. In addition, utilizing the RWKV model for predicting stock prices highlights its potential in time series forecasting tasks~\cite{Stock-Prediction-Using-RWKV, dong2024dft, cao2024matcc}.

The intelligent completion of code represents a significant application of RWKV, surpassing conventional RNN models in both understanding and completing code. This superiority is attributed to enhanced accuracy derived from faster training processes, an increased number of network layers, and the implementation of specific syntax tree mappings~\cite{zhou2023code,zhou2023rwkv}. Recent advancements have incorporated multiple temporal perspectives into the model, further enhancing its capacity to interpret context more effectively while preserving linear computational complexity~\cite{dumitru2024enhancing}. Additionally, RWKV has shown impressive proficiency in moderating content, benefiting from the use of a uniquely crafted dataset and purpose-driven experiments. This effectiveness is further bolstered by leveraging a comprehensive dataset for knowledge distillation when handling large-scale tasks~\cite{suzuki2024sensorimotor, yildirim2024experimentation}.

In terms of efficiency and resource management, the GoldFinch framework proposes a linear attention mechanism that enhances the RWKV's cache production, leading to substantial advancements in reasoning abilities across extended context lengths~\cite{goldstein2024goldfinch}. To overcome the drawbacks of conventional RNNs, the RWKV-TS model delivers a robust structure specifically designed for handling time series applications~\cite{hou2024rwkv}. The TIM model integrates the RWKV architecture by incorporating causal interactions and amplifying local patterns, resulting in superior generation outcomes~\cite{wang2024temporal}.

The Retrospected Receptance Weighted Key Value (RRWKV) framework incorporates retrospective features that improve its capacity to understand long-term dependencies, thereby overcoming the challenges faced by RWKV in processing distant information~\cite{wang2023rrwkv}. In addition, the application of transfer learning techniques significantly boosts the efficiency of domain generation algorithm classifiers, which facilitates accurate distinction between non-malicious and harmful exemplars~\cite{drichel2024transfer}. In the realm of personalized news recommendation systems, the fusion of large language models with knowledge graphs significantly boosts the precision of recommendations~\cite{chen2024lkpnr}. The Decision-RWKV model, which blends decision transformers with experience replay frameworks, exhibits strong performance in reinforcement learning and decision-making tasks~\cite{dong2024optimizing}. To combat the forgetting phenomenon prevalent in language models, a strategy utilizing synthetic gradients is introduced. This method encodes prompts directly, thereby enhancing memory retention during text generation~\cite{luo2023prosg}. Leveraging the RWKV architecture, the SpikeGPT model employs event-driven sparse activations to generate natural language effectively~\cite{zhu2023spikegpt}.

\subsection{Computer Vision}
Within computer vision, RWKV models are capturing growing interest due to their ability to integrate visual information with sophisticated analytical skills. Taking advantage of the distinct characteristics of RWKV frameworks, these models are capable of considerably improving a range of tasks. These tasks include segmenting medical images, perceiving 3D point clouds, and detecting actions in real-time.

RWKV models have made significant contributions to the field of medical image processing. BSBP-RWKV, for example, excels in the segmentation of medical images, merging traditional denoising techniques with a robust network framework for high precision and efficiency~\cite{zhou2024bsbp}. Likewise, Restore-RWKV stands out in medical image restoration by implementing an innovative omnidirectional token translation layer and reconstructive attention mechanism, which successfully capture spatial relationships in 2D images~\cite{yang2024restore}. Additionally, RWKV-based Multi-Instance Learning (MIL) has achieved substantial success in classifying high-resolution Whole Slide Images (WSI), proficiently learning global feature representations~\cite{ji2023rnn}.

In the domain of 3D point cloud perception, RWKV models demonstrate considerable proficiency. PointRWKV employs linear complexity and enhanced multi-head matrix states to process point cloud data, greatly boosting the ability to capture local geometric features~\cite{he2024pointrwkv}. Moreover, the LION framework employs a linear group RNN to adeptly model long-range relationships, adapting well to the nature of sparse point clouds, thereby ensuring efficient 3D object detection~\cite{liu2024lion}. OccRWKV serves as an efficient semantic occupancy network, utilizing RWKV modules for precise predictions within sparse environments~\cite{wang2024occrwkv}.

In the domain of real-time action detection, the TLS-RWKV model is noteworthy for its ability to grasp temporal patterns without sacrificing computational efficiency. This capability allows for the prompt prediction of action categories in streaming video, as detailed in~\cite{zhu2024tls}. RuleVM sets itself apart with a dual-branch framework tailored for the weakly supervised monitoring of violent content. It fuses visual features with language-image alignment to enhance both interpretability and performance as noted in~\cite{jiang2024explicit}. Additionally, the RWKV models have significant applications in image generation and denoising. For instance, the Diffusion-RWKV is pertinent to high-resolution image generation tasks, adeptly handling extensive inputs by mitigating spatial aggregation complexities~\cite{fei2024diffusion}. LineRWKV is efficient in hyperspectral image compression, employing line-based inference mechanisms to curtail memory use~\cite{valsesia2024hybrid}. In the realm of image segmentation, RWKV-SAM delivers efficient results with a hybrid backbone architecture, markedly outperforming traditional transformer methods~\cite{yuan2024mamba}.

Regarding high-resolution image processing, Vision-RWKV (VRWKV) demonstrates the potential to handle high-resolution images by reducing spatial aggregation complexity, surpassing the performance of Vision Transformers (ViT)~\cite{duan2024vision}. VisualRWKV-HD and VisualRWKV-UHD focus on processing high-resolution visual inputs, significantly enhancing performance in text-rich tasks~\cite{li2024visualrwkv}. The exploration of RWKV-based facial expression recognition technology continues to unfold its applications in affective computing~\cite{FaceRWKV}. Additionally, research on RWKV for image noise reduction provides new insights into image processing~\cite{rwkv-denoise}. An enhanced transformer model, RWKV-IR, is introduced to improve image restoration with advanced linear attention mechanisms~\cite{du2024exploring}.

The LSTM CrossRWKV (LCR) framework employs a Cross RWKV gate along with an improved LSTM mechanism to adeptly capture both spatial and temporal aspects pertinent to video analysis~\cite{yin2024video}. For image-text interactions, RWKV-CLIP stands out as the pioneer RWKV-integrated visual-language representation model, merging the parallel training efficiency of transformers with the inference efficacy of RNNs~\cite{gu2024rwkv}. Studies involving Spiking Neural Networks (SNNs) underscore their capabilities in generating images~\cite{yang2024sdit}. Moreover, approaches for predicting multimodal pedestrian trajectories utilize conditional variational autoencoders to adeptly account for the diversity and uncertainty in human motion~\cite{xu2023social}. StyleRWKV~\cite{dai2024stylerwkv} leverages RWKV models for efficient style transfer, introducing Re-WKV attention, a Deform-Shifting layer, and Skip Scanning to enhance global and local context while maintaining low memory usage and linear time complexity.

\subsection{Audio and Music Tasks}
In music generation, the Music-Diff architecture enhances sample diversity by using a joint probability diffusion model, overcoming the limitations of traditional language models in generative tasks~\cite{liu2024perturbing}. RWKV has extended its capabilities to MIDI music generation~\cite{Android-RWKV-MIDI, RWKV-v4-MIDI} and has been applied in music genre classification~\cite{Music-Genre-Classification-RWKV}. Its creative versatility is further demonstrated in a mini-game that integrates Stable Diffusion with the RWKV music model~\cite{Procedural-Purgatory}.

In the field of speech recognition, RWKV has made notable advancements. RWKV-based semantic voice activity detection (VAD) shows enhanced resilience to noise interference~\cite{zuo2023advancing}. At the same time, the RWKV-Transducer model has achieved competitive accuracy in streaming Automatic Speech Recognition (ASR), optimizing both latency and memory consumption~\cite{an2023exploring, RWKV-ASR}. Additionally, RWKV-SpeechChat uses the frozen 3B RWKV model to support real-time conversational tasks, including ASR, speech translation, and speech question answering~\cite{RWKV-SpeechChat}.  RWKV has also been successfully applied in audio pattern recognition~\cite{AudioRWKV}.

\subsection{Web Application }
The exceptional flexibility and easy accessibility of RWKV models have significantly driven the development of a wide variety of web-based applications. This development has, in turn, considerably enhanced user engagement with advanced AI technologies. By merging user-friendly interfaces with robust back-end frameworks, these implementations have effectively connected modern models with end-users. As a result, they have simplified seamless interactions and facilitated efficient deployment across numerous platforms.

Implementations based on web technologies have notably increased the usability and interactivity of RWKV models~\cite{MoZi-RWKV, RWKV_3B_7B_Webui_GPT-SoVITS, RWKV_QQBot_BackEnd}. As an illustration, the RWKV v4 version accessible via web browsers features a user-friendly interface, facilitating easy model access and use~\cite{rwkv-v4-web}. The role-playing web UI, created with Gradio, exemplifies the adaptability of RWKV for interactive applications~\cite{RWKV_Role_Playing}. In addition, projects focusing on both local and server-side web interfaces offer efficient means to incorporate RWKV~\cite{ai00_server,web-rwkv-axum}.

For improving user engagement, the ChatRWKV web chat interface facilitates real-time conversations between users and the model~\cite{WebChatRWKVstic,chatux-server-rwkv,ChatRWKV-flask-api,ChatRWKV-webui}. Furthermore, derived from the rwkv.cpp initiative and combined with GPT-SoVITS for novel reading, this highlights the broad applicability of RWKV in multimedia contexts~\cite{rwkv.cpp_webui_GPT-SoVITS,rwkv.cpp_webui_Macos}. The RWKV Flask service provides a minimalistic solution for deploying applications, and the Vue.js-based user interface for the RWKV Role Playing API ensures a refined user interaction experience~\cite{RWKV-api,RWKV_Role_Playing_UI}.

\subsection{Evaluation of RWKV Models}
Thorough evaluation of RWKV models is essential to fully grasp their performance and efficacy across a broad range of natural language processing benchmarks. The assessment process employs an array of metrics and methodologies crafted to evaluate both qualitative and quantitative dimensions of the model outcomes. As RWKV models have advanced, a number of evaluation methods have been developed to assess their performance. In this research, we have assembled 17 benchmark tests, detailed in Table \ref{tb-evaluation}, each highlighting distinct facets and criteria for evaluation.

\input{Tables/tb-evaluation}

Several studies have focused on evaluating the performance of RWKV in long-sequence modeling and in-context learning~\cite{akyurek2024context, leeattention, yuan2024kv}. The LooGLE benchmark evaluated the RWKV model's performance on tasks that involve long dependencies, noting that it did not excel in handling complex tasks with extended dependencies~\cite{li2023loogle}. Meanwhile, the RULER benchmark identified a noticeable drop in the model's effectiveness as the length of in-context inputs increased, a challenge particularly evident in understanding within context~\cite{hsieh2024ruler}. Furthermore, the comprehensive S3EVAL evaluation suite was utilized to examine RWKV's capabilities in long in-context tasks, exposing performance limitations when managing extremely lengthy in-context scenarios with the current versions of RWKV models~\cite{lei2023s3eval}.

The MAGNIFICO evaluation framework and the MANGO benchmark were utilized to assess RWKV's performance on tasks requiring comprehension of new vocabulary, multiple reinterpretations, and spatial reasoning capabilities. While RWKV showed promise in managing complex tasks, it revealed notable deficiencies in accurately executing spatial reasoning and swiftly adapting to rapidly changing contexts~\cite{patel2023magnifico, ding2024mango}. To measure the model's efficiency in real-world scenarios, the SuperCLUE benchmark was introduced, highlighting the inadequacy of traditional accuracy measures in truly capturing user requirements~\cite{xu2023superclue}. Furthermore, the ZhuJiu benchmark, employing a multidimensional evaluation approach, carried out an extensive analysis of RWKV's capabilities in knowledge graph and in-context learning tasks, with a particular emphasis on its performance in both Chinese and English language tasks~\cite{zhang2023zhujiu}.

In the evaluation of model robustness against malicious injection attacks, the BIPIA benchmark was proposed to assess the ability of LLMs to resist external malicious instructions~\cite{yi2023benchmarking}. Similarly, the CoDI-Eval benchmark was used to evaluate LLM performance in executing constrained instructions, revealing gaps between open-source and commercial LLMs in following instructions~\cite{chen2024benchmarking}. A voting-theory framework was applied to LLM evaluation, enhancing robustness in complex tasks by transforming tasks into voting problems~\cite{lanctot2023evaluating}.

The CMATH dataset evaluated the capability of LLMs in tackling mathematical problems, demonstrating that GPT-4 achieved higher accuracy than RWKV, especially when faced with distracting data~\cite{wei2023cmath}. The Head-to-Tail benchmark served to assess LLMs on knowledge graph Q\&A tasks, identifying RWKV's weaknesses in handling infrequent information, notably in knowledge retention and cognitive reasoning~\cite{sun2023head}. Research comparing recursive models and transformer models indicated that RWKV surpassed transformers in specific tasks, particularly those necessitating recursive processing~\cite{michaelov2024revenge}.

Research has illustrated RWKV's impressive ability to perform in-context learning tasks effectively, revealing its capacity to understand intricate data relationships within the context itself, eliminating the dependence on additional optimization tools~\cite{grmveliki}. In assessments concerning brain-machine interfaces and edge device applications, analyses comparing RWKV with GRU and Transformer models demonstrated that RWKV excels in reasoning speed and scalability. This makes it a more fitting choice for handling extensive data and models on a large scale~\cite{zhou2024benchmarking}.

Efforts to adapt transformer interpretability techniques to RNNs (notably RWKV) have been investigated, with suggestions to improve model performance by optimizing RNN state compression characteristics~\cite{paulo2024does}. Assessments of large models in regression tasks revealed that RWKV demonstrates a superior in-context learning capacity in multi-task contexts compared to conventional supervised models~\cite{vacareanu2024words}. Analyses of RWKV’s capability in long sequence modeling exposed constraints when managing very lengthy sequences~\cite{huang2024well}. The peer-reviewed LLM framework addresses biases in traditional evaluations by employing multiple LLM "reviewers" to provide a more balanced assessment, emphasizing the necessity for diverse perspectives~\cite{chu2024pre}. In examining temporal generalization, the dynamic generation benchmark proposes a methodology to highlight performance decline with time-related changes, indicating a need for enhancements in current update mechanisms to better accommodate evolving information~\cite{zhu2024evaluating}. The LongICLBench benchmark concentrated on RWKV's outcomes in challenging label classification tasks, identifying shortcomings when facing extensive label sets and long inputs, especially in complex classifications~\cite{li2024long}. Additionally, various evaluations have been performed on RWKV's abilities in projects like the Long Range Arena~\cite{rwkv-long-range-arena}, LongBench\_RWKV~\cite{LongBench_RWKV}, MT\_BENCH\_RWKV~\cite{MT_BENCH_RWKV}, and Compare\_LLMs~\cite{compare_llms}.

\subsection{Others}
The implementation of RWKV extends into several pioneering fields beyond the conventional applications in Natural Language Processing and Computer Vision, warranting significant interest. As RWKV models persistently evolve, their range of applicability continues to expand comprehensively. This part will delve into a variety of supplementary use cases, emphasizing the potential and adaptable nature of RWKV across a multitude of domains.

RWKV, when incorporated with Spiking Neural Networks (SNNs), has shown improved adversarial robustness in tasks combining vision and language~\cite{chenspiking}. The Blip2RWKV+QFormer model is tailored for dialogues involving both images and text~\cite{MiniRWKV-4}. Recent progress has seen the introduction of retrieval-augmented generation (RAG) and multimodal extensions to RWKV, broadening its functionality~\cite{RWKV-RAG}. Furthermore, VisualRWKV exemplifies its use in Visual Language Models (VLMs), leveraging a linear RNN model to deliver performance that rivals transformer-based systems~\cite{hou2024visualrwkv}.

A method for image-text alignment employing a frozen RWKV-4-World-0.4B model highlights its potential in multimodal tasks~\cite{Dlip-RWKV}. The breadth of RWKV's capabilities encompasses a wide array of tasks, including sequence classification, embedding generation, and cross-encoder applications, demonstrating its versatility in tackling various challenges in machine learning~\cite{RWKV_LM_EXT}. Moreover, RWKV has been successfully applied in the training of reward models enhanced by human feedback within the context of reinforcement learning~\cite{rwkv-reward-enhanced}. Its applications further extend to the identification of Twitter users who are bots~\cite{Bot-Ani-RWKV-twitter-bot-detection}.

Numerous studies focus on enhancing the performance and efficiency of RWKV models. For instance, one study provides a unified perspective on attention-free layers, highlighting their sub-quadratic complexity and favourable scaling properties~\cite{zimerman2024unified}. Additionally, techniques for parameter-efficient fine-tuning of RWKV~\cite{RWKV-PEFT}, along with a quick fine-tuning package for the RWKV-PEFT project~\cite{RWKV-PEFT-Simple}, and the implementation of infinite-length training for RWKV5~\cite{RWKV5-infctxLM}, as well as improvements in decision-making tasks for lifelong robotics~\cite{DecisionRWKV}, all demonstrate the model's capabilities. Separately,  RWKV has found applications in gaming, where it enhances the user experience through dynamic decision-making processes~\cite{RWKV-nonogram}. In the field of industrial time series forecasting, the RWKV model has demonstrated its potential in handling high-dimensional data~\cite{RWKV_RUL}. Other projects demonstrate the integration of RWKV with various frameworks and tools, highlighting its broad application potential~\cite{TrainChatGalRWKV, tinyrwkv, simple_rwkv}.

The integration of RWKV with an array of frameworks and tools greatly expands its range of applications. A comprehensive tool for automated RWKV management and deployment provides an interface that is compatible with the OpenAI API~\cite{RWKV-Runner}. Moreover, RWKV has been successfully integrated into Node.js~\cite{llama-node}, while another initiative showcases its modular integration capabilities with the Godot engine~\cite{godot-rwkv}. In addition, models from the RWKV Raven/Pile/PilePlus series can be converted into the HF format, facilitating full fine-tuning with Alpaca~\cite{HF-For-RWKVRaven-Alpaca}.

In the field of reinforcement learning, the utilization of RWKV is on the rise. For instance, studies examine RWKV's role in training RLHF models~\cite{rwkv_reward} and its potential applications in decision-making tasks with limitless context lengths~\cite{Decision-RWKV}. The RWKV reinforcement learning toolkit boosts capabilities in this area~\cite{RWKV-LM-RLHF} and has demonstrated impressive results in testing RWKV v6 state tuning with 4-bit quantization~\cite{RWKV-LM-State-4bit-Orpo}. Additionally, MARWKV, which is based on RWKV, achieves performance on par with MAT in multi-agent reinforcement learning, while providing superior computational efficiency as the number of agents grows~\cite{thapa2024modern}.

Diverse resources are accessible for those engaged in the fine-tuning and optimization of RWKV models. These encompass detailed tutorials on fine-tuning~\cite{RWKV-finetune-script, HF-For-RWKVWorld-LoraAlpaca, RWKV-LM-LoRA-ja}, along with a streamlined inference API~\cite{RWKV-Light-API}. Additionally, there are training tutorials focused specifically on the state tuning of RWKV models~\cite{RWKV-StateTuning}. For optimizing prefill and training speed, one can refer to a comprehensive tutorial employing chunked parallelism~\cite{chunkRWKV6}.

The significance of enhancing the interpretability and evaluation capabilities of RWKV models is comparable. Examining the interpretability of outliers within RWKV models showcases the progress made in this research sector~\cite{RWKV-LM-Interpretability-Research}. Visualization tools that display RWKV's internal states during inference~\cite{web-rwkv-inspector} and methods to document state transitions~\cite{rwkv_umap} facilitate a more profound comprehension of how these models operate. Moreover, converting RWKV models into TensorRT and ONNX formats increases the versatility of their deployment~\cite{rwkv-tensorrt}.

The notable performance of RWKV across various applications is worth highlighting. Notable implementations involve the utilization of the GGML library for offline operations on iOS and macOS, as detailed in~\cite{LLMFarm}, as well as prompt projects that integrate RWKV with Langchain, referenced in~\cite{RWKV_chains}. Additionally, the visualization service created for the RWKV-Ouroboros project, cited in~\cite{RWKV-Ouroboros-app}, exemplifies the vast applicability of RWKV. Moreover, deployment scripts for the automatic deployment of GGML/GGUF models on Android have been developed, as seen in~\cite{AltaeraAI}. Compression techniques specifically designed for RWKV, such as low-rank approximation, sparsity predictors, and clustering heads, have achieved significant results, enabling up to 4.95–3.8x compression with minimal accuracy loss~\cite{choe2024rwkv}. The integration of RWKV with the Llama architecture~\cite{RWKVinLLAMA} and the implementation of GPTQ quantization techniques~\cite{GPTQ-for-RWKV} further underscore the innovative progress within this area.

Transferring RWKV-LM models to the ROCm platform, as outlined in~\cite{RWKV-LM-AMD-Radeon-ROCm-hip}, employing inference through OpenGL technology~\cite{InnerDNN}, and merging RWKV with server-based applications designed for question-answering tasks~\cite{EasyChat-Server} create promising new opportunities for discovery. The PC version of ChatRWKV~\cite{ChatRWKV_PC} is equipped with a user-friendly interface, enhancing accessibility. Moreover, the exploration into generative calibration methods for RWKV, as discussed in~\cite{jiang2023generative}, focuses on strategies for enhancing model performance during context learning. In V2X environments, an RWKV-DPA~\cite{fu2024linear} estimator was proposed, combining RWKV network and data-pilot aided estimation. It outperformed other DL-based estimators under IEEE 802.11p standard with lower complexity. The availability of open datasets for RWKV~\cite{PlantFlowerDatasets} provides crucial resources supporting researchers in their work. The PCF-RWKV~\cite{li2024pcf} model leverages RWKV architecture, task-specialized LoRAs, and Multi-Agents technology to automate lifecycle inventory construction and carbon footprint calculation. The L3TC~\cite{zhang2024l3tc} method integrates RWKV models, an outlier-aware tokenizer, and high-rank reparameterization to enable efficient, low-complexity text compression.

%% file: Tables/tb-evaluation.tex
\begin{table*}[htbp]
  \centering
  \caption{Summary of RWKV model evaluation}
  \label{tb-evaluation}
    \begin{tabular}{cc}
    \toprule
    Benchmark & Focus \\
    \midrule
    BIPIA\cite{yi2023benchmarking} & Indirect prompt injection attacks \\
    CoDI-Eval\cite{chen2024benchmarking} & The capability of LLMs to respond to the constraints in instructions. \\
    CMATH\cite{wei2023cmath} & Mathematical problem solving \\
    VasE\cite{lanctot2023evaluating}  & Evaluation through voting theory framework \\
    Head-to-Tail\cite{sun2023head} & Assess the ability of LLMs to internalize facts \\
    REGBENCH\cite{akyurek2024context} & In-context Language Learning(ICLL) \\
    Freshbench\cite{zhu2024evaluating} & A dynamic assessment framework for temporal generalisation \\
    LongctxBench\cite{yuan2024kv} & Efficiency in handling long contexts \\
    LongICLBench\cite{li2024long} & Benchmark for extreme-label classification in long contexts \\
    LooGLE\cite{li2023loogle} & Benchmark for long-dependency understanding. \\
    MAGNIFICO\cite{patel2023magnifico} & Learning new interpretations in-context. \\
    MANGO\cite{ding2024mango} & Benchmark for mapping and navigation capabilities. \\
    PRE\cite{chu2024pre}   & Evaluation framework inspired by academic peer review. \\
    RULER\cite{hsieh2024ruler} & Enhanced benchmark for long-context evaluations extending the NIAH test. \\
    S3EVAL\cite{lei2023s3eval} & Flexible evaluation method using complex synthetic tasks. \\
    SuperCLUE\cite{xu2023superclue} & Comprehensive benchmark for Chinese user preferences. \\
    Zhujiu\cite{zhang2023zhujiu} & Multi-dimensional evaluation benchmark for Chinese LLMs. \\
    \bottomrule
    \end{tabular}%
  \label{tab:addlabel}%
\end{table*}%

%% file: tex/future.tex
\section{Future Directions and Challenges}\label{sec:challenges}
The RWKV model holds significant promise as an alternative to conventional architectures like Transformers and RNNs. It has shown notable effectiveness in multiple tasks. Nevertheless, the widespread implementation and use of RWKV are confronted with numerous challenges that necessitate deeper investigation and research.

\subsection{Expanding RWKV's Capabilities in Future Applications}
\subsubsection{Long sequence processing capability}
Turning towards the future prospects of long-sequence processing using RWKV reveals great promise for theoretical innovations and practical uses. Notably, a key avenue for exploration lies in refining the model's capacity to effectively capture and maintain information over exceedingly long contexts. Although RWKV’s mechanism mimicking recursion already provides a reduction in computational complexity when compared to Transformers, overcoming the issue of information decay across extended sequences is still a challenge. Advancements in memory mechanisms or enhancements to the architecture through explicit tracking of long-range dependencies might further boost its capabilities. Furthermore, coupling RWKV with methodologies like adaptive tokenization or hierarchical attention frameworks could lead to more efficient handling of ultra-long inputs.

\subsubsection{Multi-modal and Cross-modal Learning}
RWKV's distinctive structure offers promising avenues for advancements in multi-modal and cross-modal learning, although its full capabilities have not yet been thoroughly examined. Future studies might aim at developing specialized multi-modal model architectures, allowing RWKV to adeptly handle various forms of data, including text, audio, and visual content, while enhancing inter-modal linkages. In tasks involving cross-modal generation, RWKV could be refined to facilitate smoother transitions between modalities, such as the conversion of text to images or audio to subtitles. An expansion of this research might entail employing RWKV in more intricate modality mixtures, like audio-visual or tactile-language tasks, especially in contexts demanding multi-modal reasoning and decision-making. By integrating data preprocessing strategies specifically designed for different modalities or incorporating recent advances in multi-modal modeling techniques, RWKV could emerge as a pivotal asset in the field of multi-modal learning.

\subsubsection{Parameter-efficient Finetuning}
Parameter-efficient finetuning (PEFT) techniques have become popular as a means to adapt large models to specific tasks efficiently, minimizing the need for extensive computational power. Due to RWKV's modular structure and its computation that mimics recurrent methods, it is particularly well-suited for these approaches. Strategies like LoRA~\cite{hu2021lora}, adapters, and prompt tuning can be integrated smoothly into RWKV, facilitating task-specific adjustments while leaving the majority of the model's parameters unchanged. This approach not only reduces memory utilization but also speeds up finetuning on datasets specific to various domains. Future investigations might focus on crafting new PEFT strategies designed specifically for RWKV’s distinctive structure, potentially exploiting its sequential processing capabilities to enable effective transfer learning in time-series tasks or other sequential contexts.

\subsection{Security and Stability}
\subsubsection{Safety and Robustness}
Like other foundational models, RWKV may be susceptible to adversarial attacks, which can threaten its trustworthiness in safety-sensitive contexts. Although its recurrent-like structure provides benefits for handling sequential data, the model's behavior in adversarial settings or when faced with distributional shifts is not yet well understood. It is vital to guarantee the stability of RWKV's outputs against small input changes, especially as its use expands in critical areas. Future investigations could explore methods such as adversarial training to bolster RWKV's resistance to harmful manipulations.

\subsubsection{Bias and fairness}
Models built on the RWKV framework are trained using vast datasets, which frequently mirror societal biases and stereotypes present in their training data. These biases can subtly affect the model's behavior and might result in outcomes that are both unintended and unfavorable. Although notable strides have been made in dealing with fairness issues within other foundational model frameworks, focused studies on strategies to reduce biases specific to RWKV-based models are still relatively scarce. It is crucial to examine methodologies for pinpointing and mitigating these biases. Potential approaches encompass ensuring a balanced representation within training datasets, integrating fairness constraints during the fine-tuning phase, and implementing corrective adjustments post hoc.

\subsubsection{Privacy}
Concerns regarding privacy pose a considerable challenge for large models, especially in interactive applications, where they frequently process user data to generate outputs or predictions. In these situations, sensitive user information—such as personal identifiers, preferences, or behavior patterns—may inadvertently be stored by the model, elevating the risk of data misuse or unintentional exposure. To tackle these issues, it is necessary to develop strong privacy-preserving methods specifically crafted for the RWKV architecture. For example, incorporating differential privacy during training can prevent the model from memorizing or retrieving individual data points. Additionally, federated learning offers a promising solution by allowing local training on decentralized data without gathering raw user data centrally.

\subsection{Optimizing RWKV for Hardware Acceleration}
Future research endeavors should aim to increase the efficiency of RWKV-based models across various hardware platforms by tailoring them for low-power applications, such as edge devices and mobile gadgets. To achieve this, potential strategies could involve model quantization, utilizing sparse matrix algorithms, and implementing dynamic load balancing to maintain optimal performance in environments with limited resources. Additionally, creating optimization algorithms specifically for hardware like GPUs, TPUs, and custom ASICs offers a valuable opportunity. Customizing computation graphs and memory allocation methods to leverage the unique capabilities of these devices could lead to notable enhancements in both computational efficiency and processing speed.